\documentclass[conference]{IEEEtran}
\IEEEoverridecommandlockouts
\usepackage{cite}
\usepackage{amsmath,amssymb,amsfonts}
\usepackage{graphicx}
\usepackage{textcomp}
\usepackage{xcolor}

\usepackage{subcaption}
\usepackage{algorithm}
\usepackage{algpseudocode}
\usepackage{bm}
\usepackage{amsmath}
\usepackage{amssymb}
\usepackage{amsfonts}
\usepackage{amsthm}
\usepackage{adjustbox}
\usepackage{threeparttable}
\usepackage{multirow}
\usepackage[switch]{lineno}  %

\def \S  {\mathcal{S}}
\def \X {\mathcal{X}}

\DeclareMathOperator*{\argmax}{argmax}

\usepackage{pifont}
\newcommand{\xmark}{\ding{55}}%

\def\BibTeX{{\rm B\kern-.05em{\sc i\kern-.025em b}\kern-.08em T\kern-.1667em\lower.7ex\hbox{E}\kern-.125emX}}
\begin{document}

\title{Neural Architecture Search via Combinatorial Multi-Armed Bandit}

\author{
	\IEEEauthorblockN{Hanxun Huang\IEEEauthorrefmark{1}, Xingjun Ma\IEEEauthorrefmark{2}, Sarah M. Erfani\IEEEauthorrefmark{1}, James Bailey\IEEEauthorrefmark{1}}\\
	\IEEEauthorblockA{\IEEEauthorrefmark{1} School of Computing and Information Systems, The University of Melbourne, Victoria, Australia 
		\\\{hanxunh\}@student.unimelb.edu.au}\\
	\IEEEauthorblockA{\IEEEauthorrefmark{2}School of Information Technology, Deakin University, Geelong, Australia}
}

\maketitle

\begin{abstract}
Neural Architecture Search (NAS) has gained significant popularity as an effective tool for designing high performance deep neural networks (DNNs). 
NAS can be performed via reinforcement learning, evolutionary algorithms, differentiable architecture search or tree-search methods. 
While significant progress has been made for both reinforcement learning and differentiable architecture search, tree-search methods have so far failed to achieve comparable accuracy or search efficiency.
In this paper, we formulate NAS as a Combinatorial Multi-Armed Bandit (CMAB) problem (CMAB-NAS). This allows the decomposition of a large search space into smaller blocks where tree-search methods can be applied more effectively and efficiently.
We further leverage a tree-based method called Nested Monte-Carlo Search to tackle the CMAB-NAS problem.
On CIFAR-10, our approach discovers a cell structure that achieves a low error rate that is comparable to the state-of-the-art, using only 0.58 GPU days, which is 20 times faster than current tree-search methods. Moreover, the discovered structure transfers well to large-scale datasets such as ImageNet.
\end{abstract}


\section{Introduction}
\par
Deep neural networks (DNNs) have demonstrated superior performance on a wide range of complex learning problems such as image classification \cite{imagenet_cvpr09, krizhevsky2009learning} and object detection \cite{imagenet_cvpr09, ms_coco_2014}.
However, manual design of DNN architectures for a new domain not only requires extensive domain knowledge but also demands a huge amount of time to tune the hyperparameters.
In recent years, neural architecture search (NAS) has emerged as a powerful tool for automated design of high-performance DNNs.
Given a predefined search space, NAS applies a search strategy to find the optimal DNN architecture according to certain  performance objectives.
According to the search strategy, existing NAS methods can be categorized into four types: 1) reinforcement learning, 2) evolutionary algorithms, 3) differentiable architecture search, and 4) tree-search methods. 

The reinforcement learning (policy gradient) method was the first NAS method that was able to discover architectures competitive to hand-crafted DNNs \cite{zoph2017iclr, zoph2018learning}.
However, this type of approach can easily get stuck in sub-optimal solutions \cite{pham2018efficient, sutton2000policy}.
Evolutionary algorithms are also effective methods for NAS. However, they are known to be extremely time-consuming \cite{real2019regularized, DBLP:conf/gecco/LuWBDDGB19, DBLP:journals/corr/abs-2001-01233}.
Differentiable architecture search is so far the most effective and efficient approach, albeit it may fail in certain search spaces \cite{DBLP:conf/iclr/ZelaESMBH20}.
Compared to differentiable architecture search, tree-search methods have so far failed to achieve comparable accuracy nor search efficiency~\cite{negrinho2017deeparchitect, wistuba2017finding, liu2018progressive, wang2019alphax}.
In this paper, we propose a novel tree-search method that is as competitive as state-of-the-art differentiable architecture search methods in terms of both accuracy and efficiency, and is more robust to different search spaces.

In this work, we propose to formulate NAS as a Combinatorial Multi-Armed Bandit (CMAB) problem, a variant of the Multi-Armed Bandits (MAB) problem. We denote such a formulation of NAS as CMAB-NAS.
This formulation provides a unified framework to analyze the efficiency issues of existing tree-search methods, and help identify their key bottleneck: the random sampling of candidate architectures.
To address this bottleneck, we propose to use Nested Monte-Carlo Search to solve the CMAB-NAS problem and speedup the sampling of the candidate architectures.
In summary, our key contributions are:

\begin{itemize}

    \item We propose a novel formulation of NAS as a Combinatorial Multi-Armed Bandit problem (CMAB-NAS), and use Nested Monte-Carlo Search to tackle the problem.
    Based on CMAB-NAS, we provide a unified view of the efficiency issues of existing tree-search methods.
    
    \item On CIFAR-10, our approach can discover an architecture that is of a low error rate comparable to differentiable architecture search, using only 0.58 GPU days, which is 20 times faster than state-of-the-art tree-search methods. The discovered architecture scales well to ImageNet.
    
    \item We also show that our CMAB-NAS method can perform architecture search more robustly in different search spaces than differentiable architecture search.

\end{itemize}

\section{Related Work}

\noindent\textbf{Reinforcement Learning.} Reinforcement Learning (Policy gradient) methods were the first NAS methods that were able to discover DNN architectures that can achieve similar or even lower error rates than hand-crafted networks.
Early policy gradient methods are very time-consuming, taking $2,000 \sim 22,400$ GPU days to find a good architecture \cite{zoph2017iclr, zoph2018learning}. This has been reduced to 0.45 GPU days by weight-sharing across different child networks \cite{pham2018efficient}. However, policy gradient methods can easily get stuck in local optima \cite{pham2018efficient, sutton2000policy}, could producing less optimal architectures.

\noindent\textbf{Evolutionary algorithms.} Evolutionary algorithm-based methods are known to be time-consuming, and can require 3,150 GPU days to produce the same level of error rate as reinforcement learning methods \cite{real2019regularized}. Although this search cost has been reduced to $ 4 \sim 8$ GPU days by later works \cite{DBLP:conf/gecco/LuWBDDGB19, DBLP:journals/corr/abs-2001-01233}, they are less efficient than reinforcement learning methods \cite{pham2018efficient}.

\noindent\textbf{Differentiable architecture search (DARTS).} In \cite{liu2018darts}, the authors formulates NAS as a bi-level optimization problem in a differentiable manner, which allows more efficient search using gradient descent. A number of improved variants of DARTS have also been proposed, such as P-DARTS \cite{Chen2019pdarts}, Fair DARTS \cite{chu2019fair}, MiLeNAS \cite{MiLeNAS}, and PC-DARTS \cite{DBLP:conf/iclr/XuX0CQ0X20}. The original DARTS method takes $1.5 \sim 4$ GPU days to find a state-of-the-art architecture, which was later improved to $\sim 0.3$ GPU days by \cite{Chen2019pdarts, chu2019fair, MiLeNAS}, and further to only $\sim 0.1$ GPU days by \cite{DBLP:conf/iclr/XuX0CQ0X20}. AdaptNAS improve the generalization gap between proxy dataset and target dataset for differentiable methods \cite{li2020adapting}.  While DARTS and its variants focus on searching with proxy network and searching for convolution operations, DARTS can be directly applied on the target task \cite{cai2018proxylessnas}, searching for channel dimensions \cite{wan2020fbnetv2}, densely connected blocks \cite{fang2020densely}, resource-aware architecture search \cite{Mei2020AtomNAS} or be incorporated with continual learning \cite{wu2020firefly}. DARTS and its variants are arguably the most effective and efficient NAS methods to date. However, recent work has shown that DARTS can overfit to the validation set, and thus may fail in certain search spaces \cite{DBLP:conf/iclr/ZelaESMBH20}. \cite{Shu2020Understanding, zhou2020theory} have found that DARTS favours architectures that are of fast convergence rates, which may decrease its generalization performance. These shortcomings of DARTS motivate us to explore other types of search strategies, i.e., tree-search methods. In other words, we explore whether tree-search methods can be made as effective and efficient as DARTS methods, and at the same time, more robust to different search spaces.

\noindent\textbf{Tree-search methods.} Although tree-search methods have also been proposed for NAS \cite{liu2018progressive, wang2019alphax} and even be used to search for adversarially robust architectures \cite{chen2020anti}, they often suffer from significant efficiency issues. Even state-of-the-art tree-search methods can take $12 \sim 225$ GPU days \cite{liu2018progressive, wang2019alphax} to find a good architecture on CIFAR-10.
In this paper, we aim to improve tree-search methods under a more unified framework.

\section{Combinatorial Multi-Armed Bandit for Neural Architecture Search}
In this section, we first describe the search space of NAS: target network structure, cell structure and node structure. 
We then introduce the Combinatorial Multi-Armed Bandit (CMAB) problem and our formulation of NAS as a CMAB problem (CMAB-NAS).
Finally, we propose an efficient tree-search approach to solve the CMAB-NAS problem.

\begin{figure}[!ht]
    \centering
    \includegraphics[width=0.7\linewidth]{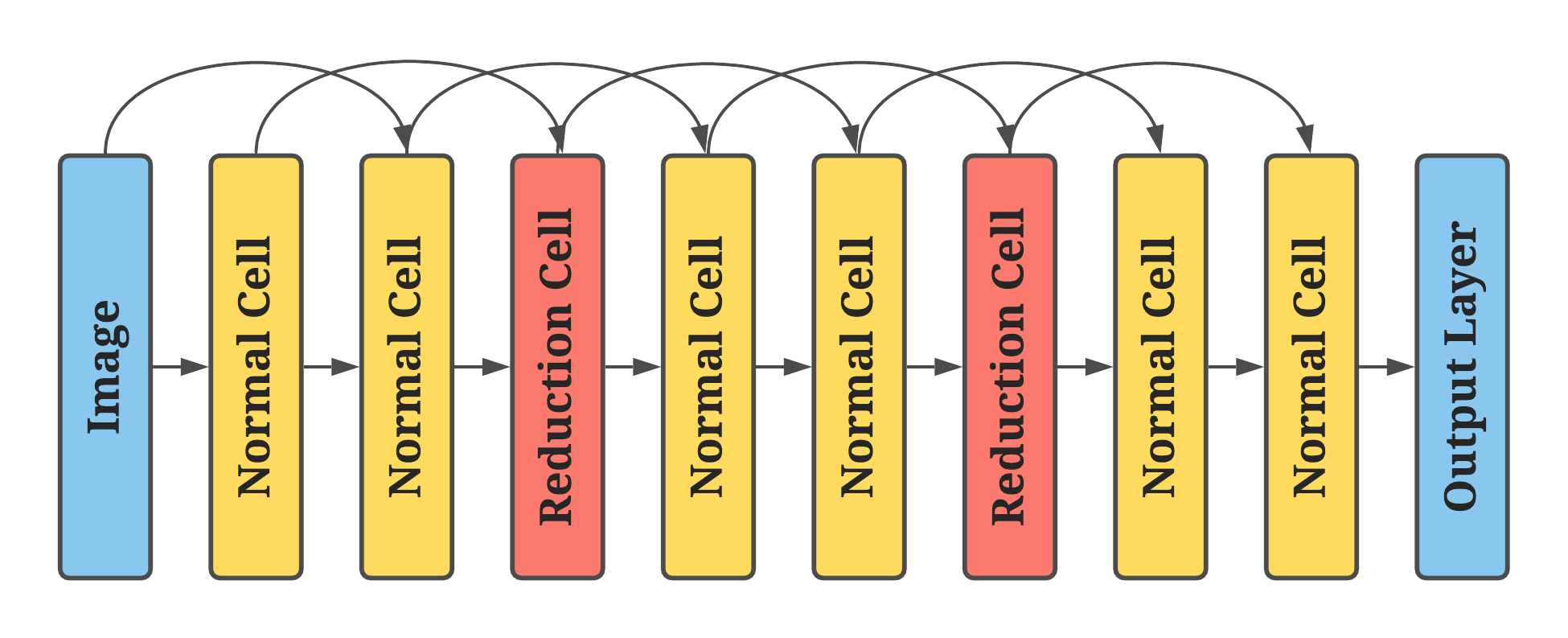}
    \caption{Architecture of the proxy network with 8 cells.}
    \label{fig:network_arc}
    \vskip -0.15in
\end{figure}

\noindent\textbf{Network structure.} As illustrated in Figure \ref{fig:network_arc}, the target network consists of several stacked convolutional cells, and the goal of NAS is to search for the optimal \emph{cell} structure of the target network.
Following previous work \cite{liu2018darts}, we perform architecture search for both the normal convolution cell ($Cell_{N}$) and the reduction convolution cell ($Cell_{R}$), that is, $(Cell_{N}, Cell_{R})$. 
The search is performed on a proxy-network, which is stacked by cells. We use the weight-sharing proxy-network \cite{pham2018efficient}, and consider the same basic structure and operations for the proxy network as in \cite{liu2018darts}, except that we removed the zero operation which is not a frequent choice in the previous work.

\begin{figure}[!ht]
	\begin{center}
        \includegraphics[width=0.95\linewidth]{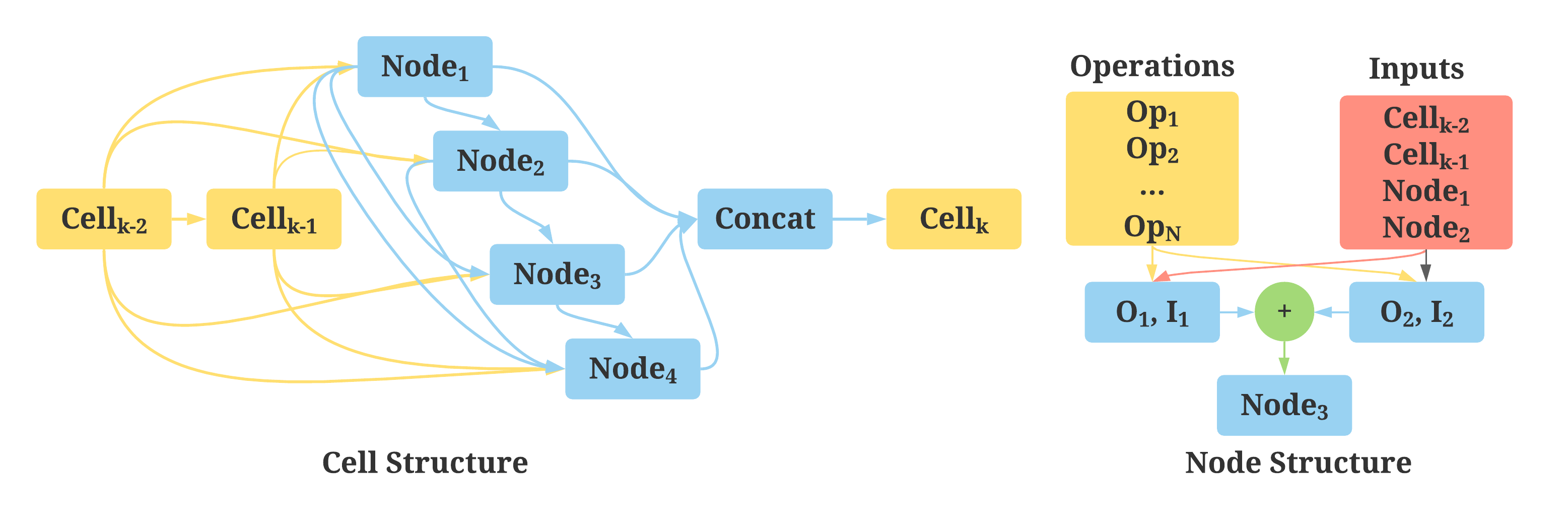}
    \end{center}
    \caption{Structures of a cell $C_k$ and a $Node_3$}
    \label{fig:cmab_search_space}
    \vskip -0.15in
\end{figure}
\begin{table}[!ht]
\caption{Basic operations in a node.}
\vskip -0.1in
\label{tab:search_space_ops}
\begin{center}
\begin{tabular}{lc}
\hline
\textbf{Operation} & \textbf{Kernel Size} \\ \hline
Skip Connection & - \\ 
Separable Convolution & $3\times3$ \\ 
Separable Convolution & $5\times5$ \\ 
Dilated Convolution & $3\times3$ \\ 
Dilated Convolution & $5\times5$ \\ 
Max Pooling & $3\times3$ \\ 
Avg Pooling & $3\times3$ \\ \hline
\end{tabular}
\end{center}
\vskip -0.1in
\end{table}

\noindent\textbf{Cell structure.} Figure \ref{fig:cmab_search_space} illustrates the cell structure to be searched, i.e., a directed acyclic graph with 4 nodes and a concatenation operation. Each node takes inputs from the two previous cells (i.e. \emph{cell inputs} $C_{k-1}$ and $C_{k-2}$), and also all its predecessor nodes (i.e. \emph{node inputs}). Note that, the first node ``Node$_1$" does not have any node input as it does not have any predecessor nodes. The outputs of all four nodes in the current cell are then concatenated to form the \emph{cell output} $C_{k}$.

\noindent\textbf{Node structure.} We denote a node by $\X_i$, which consists of two operations on two inputs, i.e., $\X_i=(I_1, O_1, I_2, O_2)$, with $I$ denoting the input and $O$ the operation.
As illustrated in Figure \ref{fig:cmab_search_space}, input $I$ can be selected from both the cell inputs and the node inputs, while operation $O$ can be selected from a set of predefined operations. 
Here, we consider the 7 basic operations defined in Table \ref{tab:search_space_ops}. 
The outputs of the two operations (i.e. $O_1$ and $O_2$) in a node are then combined into a single \emph{node output} via an element-wise addition.

\noindent\textbf{Search space size.} We denote the search space of a cell by $\S$, the total number of nodes in a single cell by $N$, and the total types of operations by $M$. We reduce the original search space $\S$ by removing the duplicates (i.e. $(I_1, O_1)+(I_2, O_2)$ represents the same node as $(I_2, O_2)+(I_1, O_1)$). This can reduce the size of the search space to $|\S|$ in Equation \eqref{search_space}, which is basically the pair of 2 distinct input-pair combinations plus the combinations where the two operations are identical, for example, $(I_1,O_1, I_1,O_1)$. In our setting, $N$=4 and $M$=7, and $|\S| \approx {6.2 \times10^9}$. The total size of our search space is $|\S|^2$ since we are searching for both the normal and the reduction cells (i.e. $(Cell_{N}, Cell_{R})$). 

\begin{equation}
\label{search_space}
|\S|= \prod\limits_{i=1}^{N}{(i+1)M\choose2} + (i+1)M
\end{equation}

\subsection{CMAB Formulation of NAS}
\label{sec:problem_formulation}
The formal definition of CMAB has been presented in previous works as a variant of Multi-Armed Bandits (MAB) \cite{chen2013combinatorial, ontanon2013combinatorial}. MAB is a classical reinforcement learning problem. Given a finite amount of resources that can be allocated for a number of choices with unknown reward distribution, 
the goal of MAB is to maximize the expected reward. 
The choices for MAB are referred to as arms. 
A solution needs to balance the exploration of the arms with unknown rewards and the exploitation of the arms with known high rewards. 
CMAB relaxes one large MAB (referred to as a global MAB) into $n$ smaller size MABs (referred to as local MABs). 
The \textit{Naive Assumption} \cite{ontanon2013combinatorial} (also known as \textit{Monotonicity Assumption} \cite{chen2013combinatorial}) allows the exploration of the global MAB through local MABs. We formulate NAS as a CMAB problem as follows:
\begin{itemize}
    \item Denote a cell (or a set of $N$ nodes) by $\X = \{\X_1, \cdots, \X_{N}\}$ with each node $\X_i$ has a search subspace with size of $|\S_i|$ different architectural choices, then each choice is called a \emph{local-arm}. The choice for all $N$ nodes in $\X$ forms a valid cell, and is called a \emph{global-arm}.
    \item The reward distribution $\mu:\X \rightarrow R$ over each cell $\X$ is unknown until the cell is determined.
\end{itemize}
A set of local-arms forming a global-arm in CMAB corresponds to a set of nodes forming a cell (i.e. $\X$) in NAS. 
The goal of NAS is to find optimal cell structures $(Cell_N, Cell_R)$ that leads to the best performing child network, or from the CMAB perspective, to find a valid global-arm that optimizes the expected reward. 
Accordingly, we can define the following local and global MAB problems for NAS.
\begin{itemize}
    \item Local MAB: Each node $\X_i \in \X$ defines a local $MAB_i$, which selects the pairs of inputs and operations for node $\X_i$.
    \item Global MAB: $MAB_g$, which considers the whole CMAB problem as a single MAB. $MAB_g$ selects the pairs of inputs and operations for the entire cell (i.e. for both $Cell_N$ and $Cell_R$).
\end{itemize}
As the global-arm can be formed by a combination of local-arms, a \emph{naive assumption} \cite{ontanon2013combinatorial} can be made between the global and the local MABs: the global reward $\mu_g$ for $MAB_g$ can be approximated by the sum of the local rewards $\mu_i$ for each $MAB_i$, and the local reward $\mu_i$ only depends on the choices made for $MAB_i$. This assumption allows $MAB_g$ to be optimized via the optimization of each $MAB_i$.
In other words, the reward function for a cell can be approximated by the sum of rewards for its all $N$ nodes:
\begin{equation} \label{navie_assumption}
\mu_g(\X) \approx \sum_{i=1}^{N}\mu_i(\X_i)
\end{equation}
This assumption allows the decomposition of the cell structure search problem into a set of smaller problems on the node structures.

\noindent\textbf{Necessity of the naive assumption.}
It may be possible to treat each global-arm as a complex local-arm in a single MAB. Then, the cell search problem will become a traditional MAB that directly searches the cell structure without being decomposed into local nodes. However, due to the huge size of the cell search space, it is extremely difficult to apply traditional approaches, where all unplayed arms need to be explored at least once. In contrast, with the naive assumption, there is no need to expand the entire cell search space rather than the local MABs. The number of arms in a local MAB$_i$ can be calculated using Equation \eqref{search_space} by taking a specific node $i$. For example, under the current search settings (eg. $N$=4, $M$=7), the first local MAB (eg. MAB$_1$) only consists of 105 (eg. ${(1+1)M\choose2} + (1+1)M$) different arms. Therefore, the CMAB with naive assumption approach is a more practical choice. We denote this formulation of NAS as \emph{CMAB-NAS}.

\noindent\textbf{Two objectives of CMAB-NAS.}
In order to obtain a higher reward (validation accuracy), the proxy network needs to update its weights through training. During the search process, the proxy network can be trained via training of the sampled child networks with weight-sharing \cite{pham2018efficient}. To better train the proxy network, we need to sample a batch of \emph{promising} child networks that have high rewards, as low reward child networks are less helpful or may even harm the next iteration's performance.
Taking this into consideration, there will be two objectives for CMAB-NAS: 1) selection of the promising child networks for training the proxy network, and 2) selection of the best performing child network for higher reward. We use regret to define the two objectives. Suppose we use selection policies $P$ and $Q$ to select promising child networks and the best performing child network respectively, the two regrets that define the above two objectives of CMAB-NAS are:
\begin{linenomath*}
\postdisplaypenalty=0
\begin{align} 
\label{eq:regrets}
\textit{cumulative regret: }& R_c = \sum_{t=1}^{T} \mu_g^* - \mu_{P(t)} \\ 
\textit{simple regret:  }& R_s = \mu_g^* - \mu_{Q(t)},
\end{align}
\end{linenomath*}
where, $t$ stands for the $t$-th play of the $MAB_g$ (for a total number of $T$ plays), $\mu_{P(t)}$ and $\mu_{Q(t)}$ denote the rewards of the $t$-th play with respect to selection policies $P$ and $Q$ respectively, and $\mu_g^*$ denotes the optimal global reward.
If the classification accuracy is used as the reward, then $\mu_g^*$ is the 100\% accuracy and the regrets become the error rates.
The \emph{cumulative regret} for a total of $T$ plays defines the objective to select promising (low error rates) child networks for training, while the \emph{simple regret} for one play of $MAB_g$ defines the objective to select the best child network for final evaluations.

\section{Proposed Tree-search Solution for CMAB-NAS} \label{sec:problem_solution}
There already exist several CMAB sampling policies, such as \emph{CUCB} \cite{chen2013combinatorial} and \emph{Naive Sampling} ($NS$) \cite{ontanon2013combinatorial}. However, they cannot be directly applied to CMAB-NAS.
Although the guaranteed logarithm cumulative regret bound of CUCB fits well to the CMAB-NAS objectives, CUCB requires an $(\alpha , \beta)$-approximation oracle, which in NAS means that we will need an additional performance prediction model to produce the approximation.
On the other hand, $NS$ has a linear cumulative regret bound, which is worse than $\emph{CUCB}$. Considering the huge amount of search space in NAS, it is not optimal for CMAB-NAS neither. 
Moreover, the two policies consider local MABs separately while ignoring the correlations between different local MABs. 
However, in NAS, some combinations of operations may perform better. For example, skip connections work well only when there are enough convolution operations, if each local MAB makes independent selections, the resulting cell structure may end up with only skip connections.
Alternatively, we propose to use the Nested Monte-Carlo Search (NMCS) \cite{cazenave2009nested} to optimize the \emph{cumulative regret}, and the top-k selection strategy to optimize the \emph{simple regret}. NMCS considers each local MAB in a contextual setting (aware of other local MABs' selections) and optimizes each local MAB to approximate the reward of the global MAB based on the \emph{naive assumption}.

\subsection{NMCS-guided Architecture Search for CMAB-NAS}
\label{sec:uct}
We adapted the Nested Monte-Carlo Search (NMCS) to our problem by modifying the sampling strategy. The original NMCS uses random sampling \cite{cazenave2009nested}, here we use UCB sampling for our problem. The UCB has been proposed to solve traditional MAB problems \cite{kocsis2006bandit}. Given a local MAB$_i$ which is a traditional MAB problem, the UCB selection strategy can be defined as:
\begin{equation} \label{ucb}
\text{UCB: } \argmax_{arm_{j} \in \X_{i}} \; \overline{\mu}(\X_{i}, arm_j) + \alpha\sqrt{\frac{2 \ln n_i}{n_j}}
\end{equation}
where, $\overline{\mu}(\X_{i}, arm_j)$ is the average reward of local-arm $arm_j$ (i.e. the architectural choice for node $\X_{i}$) in local $MAB_i$, 
$n_i$ is the number of times local $MAB_i$ has been played,
$n_j$ is the number of times a local-arm $arm_j$ has been selected until the current search iteration, and $\alpha$ is the parameter balancing the trade-off between exploration (i.e. $\sqrt{\frac{2 \ln n_i}{n_j}}$) and exploitation (i.e. $\overline{\mu}(\X_{i}, arm_{j})$).
The adapted NMCS maintains a tree structure for each possible architecture of a node and its reward distribution. The path from the root to a leaf in the tree structure defines the exact inputs and operations of a valid cell structure. Initially, the tree has unknown rewards, which can be iteratively estimated by a \emph{simulation} (Algorithm \ref{alg:uctsimulation}) process (will be explained shortly).
The sampled child architecture is evaluated on the validation set. The classification accuracy is used as the reward and updated using \emph{backprop}. We assume each node contributes equally to the final reward. Therefore, when updating the reward for each $arm_j$ in $\X_{i}$, we accumulatively add $Reward / 2N$ to its total rewards and increment the $n_i, n_j$ by one. 
Note, during sampling, we use the average reward (eg. $\overline{\mu}(\X_{i}, arm_{j})$).
The overall search procedure is described in Algorithm \ref{alg:CMAB-NAS}-\ref{alg:uct_search_best}. 
At a high level, it is an iteratively applied two-step process: 1) sample a batch of promising networks for training the proxy network (via the UCB selection), and 2) select the best child network as a candidate for the final evaluation (via the Top-k selection). 
This two-step process corresponds to the optimization of the two objectives of CMAB-NAS as follows.

\begin{algorithm}[ht!]
\caption{NMCS-guided CMAB-NAS}
\label{alg:CMAB-NAS}
\begin{algorithmic}[1]
\State {\bfseries Input:} Proxy-Network $S$, Epochs $E$, $B$, $\alpha$
\State $BestChild$ = None, $R_{max}$ = $0$
\For{$i=1$ {\bfseries to} $E$} 
    \State Initialize Tree $Tr$, $Candidates$ = empty
    \For{$j=1$ {\bfseries to} $B$} \Comment{Objective 1}
        \State Simulation$(Tr)$ 
        \State $Child_j$ = Sample$(Tr, \alpha)$ \Comment{UCB sampling}
        \State $Candidates \xleftarrow[]{} Candidates \cup Child_j$
    \EndFor
    \State Train$(S, Candidates)$
    \State $R$ = Eval$(S, Candidates)$
    \State Backprop$(Tr, R)$ \Comment{Update rewards for local MABs}
    \State $BestChild_i$ = SearchBest$(Tr)$ \Comment{Objective 2}
    \State $R_i$ = Eval$(S, BestChild_i)$
    \If{$R_i>R_{max}$}
        \State $BestChild=BestChild_i, R_{max}=R_i$
    \EndIf
\EndFor
\State {\bfseries Output:} $BestChild$
\end{algorithmic}
\end{algorithm}

\label{sec:aux_algorithms}
\begin{algorithm}[ht!]
   \caption{Sample}
   \label{alg:uctsample}
\begin{algorithmic}[1]
   \State {\bfseries Input:} Root Node $r$, $\alpha$
   \State Initialize $V = empty$
   \Repeat
   \If{$r$ is $not$ fully explored} 
   \State $r$ = $Explore(r)$ \Comment{Select the unexplored arm}
   \Else
   \State $r = UCB(r, \alpha)$
   \EndIf
   \State $V \xleftarrow[]{} V \cup r$
   \Until{$V$ is $valid$ cell structure}
   \State {\bfseries Output:} $V$
\end{algorithmic}
\end{algorithm}

\begin{algorithm}[h!]
   \caption{Simulation}
   \label{alg:uctsimulation}
\begin{algorithmic}[1]
   \State {\bfseries Input:} Proxy network $S$, Tree $Tr$, Limit $L$, $\alpha$
   \For{$i=1$ {\bfseries to} $L$}
   \State $Child$ = Sample$(Tr, \alpha)$
   \State $R$ = Eval$(S, Child)$
   \State Backprop$(Tr, R)$
   \EndFor
\end{algorithmic}
\end{algorithm}

\begin{algorithm}[h!]
   \caption{Search Best}
   \label{alg:uct_search_best}
\begin{algorithmic}[1]
   \State {\bfseries Input:} Root Node $r$, Limit $L$, $\alpha$
   \State Initialize $V = empty$
   \Repeat
   \State Simulation$(Tr, L, \alpha)$
   \State $r$ = Policy$(r)$
   \State $V \xleftarrow[]{} V \cup r$
   \Until{$V$ is $valid$ cell structure}
   \State {\bfseries Output:} $V$
\end{algorithmic}
\end{algorithm}

\noindent\textbf{Objective 1: NMCS-guided child network sampling.}
This step minimizes the \emph{cumulative regret} of CMAB-NAS,  corresponds to line 5 to 10 in Algorithm \ref{alg:CMAB-NAS}.
In each epoch (for total $E$ epochs), the tree $Tr$ interacts with a particular snapshot of the proxy network to reinitialize the reward distribution, and sample a new set of promising child networks. In the sampling process, the algorithm visits a particular node ($MAB_i$) in the tree and expands a ``new" arm if that particular node is not fully explored. Otherwise, it selects an arm following UCB in Equation \eqref{ucb}. This process is repeated until it reaches the leaf node. The selected arms from root to leaf form a valid cell structure, which further forms a valid child network. Each valid architectures is trained with one optimization (gradient) step with a randomly sampled batch of data, then evaluated on the validation set, and the reward is backpropagated to the tree $Tr$. This interaction process is called \emph{simulation}. This simulation process will repeat for $B$ times to obtain a ``\emph{ChildSet}" of $B$ child networks, which are then used to train the proxy network.

\noindent\textbf{Objective 2: Selection policy for the final child network.}
This step minimizes the \emph{simple regret} of CMAB-NAS and corresponds to line 13 in Algorithm \ref{alg:CMAB-NAS}.
After the NMCS-guided simulation and training of the proxy network, the reward distribution of the child networks can be obtained via evaluations on the validation set, along with previous simulations.
The selection of the final child network is done by the ``\emph{SearchBest}'' described in Algorithm \ref{alg:uct_search_best}.
Specifically, we use \emph{Top-k selection} (eg. $k=1$) to minimize the \emph{simple regret} (Equation \eqref{eq:regrets}).
The maximum reward can be achieved by exploiting the best performing arm. Recall the \emph{naive assumption} states that the global reward $\mu_g$ for $MAB_g$ can be approximated as the sum of rewards $\mu_i$ for $MAB_i$. It is guaranteed that if the reward of each node in set \(\mathcal{X}\) is greater than each node in set \(\mathcal{X}^\prime\), then the reward of \(\mathcal{X}\) is also greater than \(\mathcal{X}^\prime\), as mathematically described in Equation \eqref{navie_assumption_2} below. Based on this guarantee, we explore 3 different selection policies for the final child network: 1) \emph{local optimal}, 2) \emph{local suboptimal} and 3) \emph{local random}.

\begin{linenomath*}
\postdisplaypenalty=0
\begin{align} 
\label{navie_assumption_2}
\sum_{i=1}^{N}\mu_{P}(\X_{i}) >& \sum_{i=1}^{N}\mu_{Q}(\X^\prime_{i}) \\
\nonumber \text{s.t. } \; \mu_{P}(\mathcal{X}_{i}) >& \mu_{Q}(\mathcal{X}^\prime_{i}) \;  \text{ for } i = 1, \cdots, N \\
\label{optimal_policy} \text{Local Optimal: }     & \argmax_{arm_j\in\X_i} \mu_i(\X_{i}, arm_j) \\ 
\label{sub-optimal_Policy} \text{Local Suboptimal: } & \text{Second-best $arm_m$ from $\mu(x_{i})$}   \\
\text{Local Random: } & Random(\mathcal{X}_{i})\label{random_policy}
\end{align}
\end{linenomath*}

At each $MAB_i$, we refer to the policy that exploits the best and second-best arm so far as the \emph{local optimal} and \emph{local suboptimal} policy, respectively. We refer to the policy that exploits a random child network as the \emph{local random} policy. The random and local suboptimal policies serve as baselines. Since the original \emph{naive assumption} is only used for exploration of the global arm in $MAB_g$, the local policies could lose global optimality by using the \emph{naive assumption} to exploit local MABs. The proposed algorithm and \emph{local optimal} policy do not guarantee the regret bound for global $MAB_g$, but on each local $MAB_i$, the regret bound for UCB sampling is preserved. 
In practice, we find that the \emph{local optimal} policy works reasonably well for CMAB-NAS.

\subsection{A Unified View of Current Tree-search Methods}

\begin{table}[h]
\centering
\caption{Comparison with existing tree-search methods. \\\small{ C-Regret: cumulative regret. S-Regret: simple regret. P: the need of a prediction Model.}}
\label{tab: method-compare}
\begin{tabular}{cccccc}\hline
Method                                         & C-Regret    & S-Regret  & Type           & P          \\ \hline
PNAS\cite{liu2018progressive}                  & P\&Top-K        & Top-K       & SMBO  & \checkmark     \\
Wistuba\cite{wistuba2017finding}               & Random         & UCB         & MCTS  & \checkmark     \\
DeepArchitect\cite{negrinho2017deeparchitect}  & Random         & UCB         & MCTS  & \checkmark     \\
AlphaX\cite{wang2019alphax}                    & Random         & UCB         & MCTS  & \checkmark     \\
\textbf{CMAB-NAS (ours)}                                      & \textbf{UCB}           & \textbf{Top-K}       & \textbf{NMCS}  & \textbf{\xmark}          \\ \hline
\end{tabular}
\end{table}

Our CMAB-NAS formulation provides a unified framework to understand the efficiency issues of existing tree-search methods \cite{liu2018progressive, negrinho2017deeparchitect, wang2019alphax, wistuba2017finding}.
Existing tree-search methods can be converted to CMAB-NAS based on two criteria: 1) the selection strategy of the child networks for training the proxy network or equivalent processes, and 2) the selection strategy of the final child network. These two criteria correspond to the two objectives of CMAB-NAS, i.e., the \emph{cumulative regret} and the \emph{simple regret}. 
The different strategies adopted by existing tree-search methods are summarized in Table \ref{tab: method-compare}.
Our approach uses UCB sampling to optimize the \emph{cumulative regret}, while \cite{wistuba2017finding}, DeepArchitect \cite{negrinho2017deeparchitect} and AlphaX \cite{wang2019alphax} use a random strategy.
Although the proposed solution does not guarantee any global regret bound, for each local $MAB_i$, the UCB sampling can achieve the logarithm regret bound \cite{auer2002finite}. Our empirical results in Section \ref{sec:selection_policy} verify that the \emph{naive assumption} works reasonably well for NAS problems.
Combining the local UCB sampling and the \emph{naive assumption}, our approach can efficiently select a batch of promising architectures for training. 
For PNAS \cite{liu2018progressive} which uses SMBO (an improved MCTS approach), the regret bound is related to an additional prediction model and is hard to measure. It also requires additional computational cost to train the prediction model.
In fact, all existing tree-search methods rely on a prediction model or a fixed function for performance prediction.
Particularly, for AlphaX \cite{wang2019alphax} which is arguably the state-of-the-art tree-search method, the training of its Meta-DNN prediction model is time-consuming.
Our CMAB-NAS approach with UCB and Top-k selection strategies can improve tree-search methods to an efficiency level that is as competitive as state-of-the-art DARTS methods, as we will show in the experiments.

\section{Experiments}
\begin{table*}[!t]
\centering
\caption{Results on CIFAR-10. The search cost of CMAB-NAS is measured on a RTX-2080Ti. The best results for non-tree-search and tree-search methods are highlighted in \textbf{bold} separately. }
\label{tab: result-compare}
\begin{adjustbox}{max width=\textwidth}
\begin{threeparttable}
\begin{tabular}{lcccc}
\hline  
\textbf{Method} & \textbf{\begin{tabular}[c]{@{}c@{}}Test Error \\ (\%)\end{tabular}} & \textbf{\begin{tabular}[c]{@{}c@{}}Params \\ (M)\end{tabular}} & \textbf{\begin{tabular}[c]{@{}c@{}}Search Cost \\ (GPU Days)\end{tabular}} & \textbf{Type} \\ \hline

DenseNet-BC \cite{huang2017densely}                & 3.46  & 25.6  & -  & Human              \\ \hline
NASNet-A \cite{zoph2018learning} $^\star$          & 2.65  & 3.3   & 2000  & Reinforcement Learning \\
ENAS \cite{pham2018efficient} $^\star$             & 2.89  & 4.6   & 0.45  & Reinforcement Learning \\ 
AmoebaNet-A \cite{real2019regularized}             & 3.34  & 3.2   & 3150  & Evolutionary Algorithms \\
AmoebaNet-B \cite{real2019regularized}$^\star$     & 2.55  & 2.8   & 3150  & Evolutionary Algorithms \\ 
EcoNAS \cite{DBLP:journals/corr/abs-2001-01233}$^\star$ & 2.62 & 2.9 & 8   & Evolutionary Algorithms \\
DARTS (Second Order) \cite{liu2018darts}$^{\star}$ & 2.76 & 3.3 & 4 & Differentiable \\
SNAS (moderate) \cite{SNAS} $^\star$               & 2.85 & 2.8 & 1.5 & Differentiable \\
P-DARTS \cite{Chen2019pdarts} $^\star$             & \textbf{2.50} & 3.4 & \textbf{0.3} & Differentiable \\ 
Firefly \cite{wu2020firefly} $^\star$              & 2.78 & 3.3 & 1.5 & Differentiable \\ 
AdaptNAS-S (Rot-1) \cite{li2020adapting} $^\star$  & 2.59 & 3.6 & 0.5 & Differentiable \\ \hline
Wistuba\cite{wistuba2017finding} & 6.45 & - & 5 & Tree Search  \\
PNAS \cite{liu2018progressive}                      & 3.41 & 3.2 & 225 & Tree Search \\
AlphaX\cite{wang2019alphax}$^{\star \dagger}$     & 2.78 & 8.89 & 12 & Tree Search  \\
\textbf{CMAB-NAS$^\star$}                          & \textbf{2.58} & 3.80 & \textbf{0.58} & Tree Search  \\ \hline
\end{tabular}
\begin{tablenotes}
   \item[$^\star$ Cutout is used for augmentation. $\dagger$ Obtained by running open source code of the pre-trained model AlphaX-1.]
\end{tablenotes}
\end{threeparttable}
\end{adjustbox}
\end{table*}

\begin{table*}[!t]
\centering
\caption{Results of different NAS methods on ImageNet following the same setting as \cite{Chen2019pdarts,liu2018darts,wang2019alphax}. The best results for non-tree-search and tree-search methods are highlighted in \textbf{bold} separately.}
\label{tab: result-compare-imagenet}
\begin{adjustbox}{max width=\textwidth}
\begin{tabular}{lclclcc}
\hline
\multirow{2}{*}{\textbf{Method}} & \multicolumn{2}{c}{\textbf{\begin{tabular}[c]{@{}c@{}}Test Error \\ (\%)\end{tabular}}} & \multirow{2}{*}{\textbf{\begin{tabular}[c]{@{}c@{}}Params \\ (M)\end{tabular}}} & \multicolumn{1}{c}{\multirow{2}{*}{\textbf{\begin{tabular}[c]{@{}c@{}}\#Ops\end{tabular}}}} & \multirow{2}{*}{\textbf{\begin{tabular}[c]{@{}c@{}}Search Cost \\ (GPU Days)\end{tabular}}} & \multirow{2}{*}{\textbf{Type}} \\ \cline{2-3}
 & Top-1 & \multicolumn{1}{c}{Top-5} &  & \multicolumn{1}{c}{} &  &  \\ \hline
Inception-v1 \cite{inception}                   & 30.2 & 10.1 & 6.6 & 1448 & -    & Human \\ 
MobileNet \cite{mobilenet}                      & 29.4 & 10.5 & 4.2 & 569  & -    & Human \\ \hline 
NASNet-A \cite{zoph2018learning}                & 26.0 & 8.4  & 5.3 & 564  & 2000 & Reinforcement Learning \\ 
NASNet-B \cite{zoph2018learning}                & 27.2 & 8.7  & 5.3 & 488  & 2000 & Reinforcement Learning \\ 
AmoebaNet-A \cite{real2019regularized}      & 25.5 & 8.0  & 5.1 & 555  & 3150 & Evolutionary Algorithms \\ 
AmoebaNet-B \cite{real2019regularized}          & 26.0 & 8.5  & 5.3 & 555  & 3150 & Evolutionary Algorithms \\ 
AmoebaNet-C \cite{real2019regularized}          & 24.3 & 7.6  & 6.4 & 570  & 3150 & Evolutionary Algorithms \\ 
EcoNAS \cite{DBLP:journals/corr/abs-2001-01233} & 25.2 & -    & 4.3 & -    & 8    & Evolutionary Algorithms \\ 
DARTS \cite{liu2018darts}                       & 26.7 & 8.7  & 4.7 & 574  & 4    & Differentiable \\ 
SNAS (moderate) \cite{SNAS}                     & 27.3 & 9.2  & 4.3 & 522  & 1.5  & Differentiable \\
P-DARTS \cite{Chen2019pdarts}                   & 24.4 & 7.4  & 4.9 & 557  & \textbf{0.3}  & Differentiable \\
AtomNAS-A \cite{Mei2020AtomNAS}                 & 25.4 & 7.9  & 3.9 & 258  & 112  & Differentiable \\ 
AtomNAS-C \cite{Mei2020AtomNAS}                 & \textbf{24.1} & \textbf{7.3}  & 4.7 & 360  & 112  & Differentiable \\
AdaptNAS-S (Rot-1) \cite{li2020adapting}        & 24.7 & 7.6  & 5.2 & 575  & 0.5  & Differentiable \\\hline
PNAS \cite{liu2018progressive}                  & 25.8 & 8.1  & 5.1 & 588  & 225  & Tree Search \\ 
AlphaX \cite{wang2019alphax}                    & \textbf{24.5} & \textbf{7.8}  & 5.4 & 579  & 12   & Tree Search \\
\textbf{CMAB-NAS}                               & 25.8 & 8.4  & 5.3 & 619  & \textbf{0.58} & Tree Search \\ \hline
\end{tabular}
\end{adjustbox}
\end{table*}

In this section, we first compare our CMAB-NAS approach with state-of-the-art methods on both CIFAR-10 and ImageNet datasets. 
Then, we empirically verify the effectiveness of the \emph{Naive Assumption}. 
We also show that our method can perform the search more robustly in different search spaces.
We report the two standard NAS evaluation metrics: 1) the classification error of the final architecture, and 2) the search cost in GPU days.
Following previous work \cite{liu2018darts}, the architecture is searched on simple dataset CIFAR-10 \cite{krizhevsky2009learning} then transferred to other complex datasets such as ImageNet \cite{imagenet_cvpr09}.
For all our experiments, we use the same parameter settings for the proxy network and the final network as in the original DARTS method \cite{liu2018darts}. 
We use the same type of data augmentations for final network training without any additional tricks.

\noindent\textbf{Proxy Network setting.} The proxy network is stacked by 6 normal cells and 2 reduction cells with channel size 16 (as illustrated in Figure \ref{fig:network_arc}). The proxy network is trained using Stochastic Gradient Descent (SGD) optimizer with momentum 0.9, initial learning rate 0.025 and cosine scheduler \cite{loshchilov2016sgdr} without restart.

\noindent\textbf{Final Network Setting.}
For training of the final architecture, for the different dataset, we use the same hyper-parameters and techniques as in \cite{liu2018darts}. On CIFAR-10, the network consists of 20 stacked cells with a filter size of 36. The network is trained for 650 epochs using SGD with momentum 0.9, initial learning rate 0.025, and cosine scheduler \cite{loshchilov2016sgdr}. Typical techniques including cutout \cite{devries2017improved} and scheduled drop path are also applied. For ImageNet \cite{imagenet_cvpr09}, the network consists of 14 stacked cells with filter size 48 and is trained for 250 epochs using SGD with momentum 0.9, initial learning rate 0.025 and decay by 0.97 for every epoch. Colour jittering is used for data augmentation.

\noindent\textbf{NMCS Parameter setting.} 
The warm-up of proxy network is performed for 5 epochs. We search for 50 epochs. The exploration parameter $\alpha$ for UCB is set to 1.0 with decay rate of 0.95 at each epoch. We use top-1 (k=1) selection policy for the \emph{simple regret}. ChildNetSet size $B$ for Algorithm \ref{alg:CMAB-NAS} is set to 2500. The number of iterations in \emph{Simulation} is set to $L=8$ and $L=800$ for Algorithm \ref{alg:uctsimulation} and Algorithm \ref{alg:uct_search_best}, respectively. The scale up of $L$ for Algorithm \ref{alg:uct_search_best} is because a smaller number of iterations is sufficient for exploring shallow local MABs, however, it requires more iterations to ensure (before choosing the next node for the best architecture) that all local MABs are fully expanded.

\subsection{Results on CIFAR-10 and ImageNet}\label{sec:experiment_cifar10}
The results of different methods on CIFAR-10 dataset are reported in Table \ref{tab: result-compare}. 
For CMAB-NAS, here we use \emph{local optimal} policy to select the final child network.
For differentiable architecture search, we consider the original DARTS \cite{liu2018darts} method as our main competitor since we are using the same proxy network, search space and hyper-parameter setting. As can be observed, our CMAB-NAS approach is on par with DARTS in terms of error rate, but is $\sim 7\times$ (eg. 0.58 vs 4 GPU days) more efficient than DARTS and is as efficient as the P-DARTS, an accelerated version of DARTS.
This implies that our proposed backpropagation of the accumulative reward on the tree of NMCS is more efficient than the differentiable-based bilevel optimization of DARTS.
Interestingly, despite being a completely different approach, our optimization of the cumulative and the simple regrets share certain similarities with the bilevel optimization of DARTS. Our \emph{local optimal}  selection strategy for the final child network over different child nodes in the tree of NMCS has the same effect as the Softmax function of DARTS over all possible operations.
Compared with the state-of-the-art tree-search method AlphaX \cite{wang2019alphax}, our approach achieves a lower error rate, and is 20$\times$ faster. Particularly, our method can achieve a 2.58\% error rate using only 0.58 GPU days, compared to the 2.78\% error rate of AlphaX but using 12 GPU days. 
As we explained in the previous section, this is because AlphaX uses a random policy to optimize the \emph{cumulative regret}, which is less efficient than our UCB.
The training of an additional Meta-DNN network for reward prediction is another reason why AlphaX is less efficient. 
Overall, our CMAB-NAS formulation and the proposed NMCS approach with UCB and Top-1 selection policies has successfully improved tree-search methods to a performance level that is as effective and efficient as state-of-the-art reinforcement learning or DARTS methods. 
This opens up more opportunities for tree-search based NAS solutions.
The best performing cell architecture discovered by our method is illustrated in Appendix.

\subsection{Transferring to ImageNet}
\par
The results on ImageNet \cite{imagenet_cvpr09} are reported in Table \ref{tab: result-compare-imagenet}. Following previous works, here we transfer the discovered cell architectures on CIFAR-10 to ImageNet under the same setting as \cite{Chen2019pdarts,liu2018darts,wang2019alphax}. Since this is a direct architecture transfer, the efficiency does not change. 
The error rate of our CMAB-NAS is within the same range (eg. 24\% to 26\% top-1 and 7\% to 9\% top-5 test error) as state-of-the-art methods DARTS and AlphaX. Note that, certain variations may occur when transferring the architecture found on a target dataset to other datasets, as have been discussed in many previous works \cite{Chen2019pdarts, chu2019fair, liu2018progressive, liu2018darts, real2019regularized, wang2019alphax}.
For example, DARTS has lower error rate than PNAS on CIFAR-10, but has higher error rate when transfers to ImageNet. However, the transferred error rates are generally within the same range, as is also the case for our CMAB-NAS method. Overall, the results on ImageNet confirms that our CMAB-NAS approach can indeed find transferable cell architectures.

\subsection{Empirical Verification of the \emph{Naive Assumption}}
\label{sec:selection_policy}

\begin{table}[h]
\centering
\caption{Error rates of our CMAB-NAS on CIFAR-10 with different selection policies for the \emph{simple regret}. The best results are in \textbf{bold}.}
\label{tab: policy-compare-second}
\begin{adjustbox}{max width=\linewidth}
\begin{tabular}{lcccccc}
\hline
\multirow{2}{*}{\textbf{Policy}} & \multicolumn{5}{c}{\textbf{Error Rate (\%)}}        \\ 
                                & Run1  & Run2  & Run3  & Run4  & Run5  & Avg          \\ \hline
Local-optimal                   & \textbf{2.87}  & \textbf{2.91}  & \textbf{2.58}  & \textbf{2.93}  & \textbf{2.77}  & \textbf{2.81}         \\
Local-suboptimal                & 3.35  & 2.99  & 3.12  & 3.19  & 3.24  & 3.18         \\
Local-random                    & 3.63  & 3.11  & 3.10  & 3.08  & 3.26  & 3.24         \\ \hline
\end{tabular}
\end{adjustbox}
\end{table}

In the previous section, we introduced three possible selection policies for optimizing the \emph{simple regret}: local-optimal (Equation \eqref{optimal_policy}), local-suboptimal (Equation \eqref{sub-optimal_Policy}) and local-random (Equation \eqref{random_policy}).
To show the superiority of the local-optimal policy, we run our CMAB-NAS method five times with different random seeds for each of the three selection policies. The selection policy for the \emph{cumulative regret} is fixed to UCB. The results are reported in Table \ref{tab: policy-compare-second}. On average, the local-optimal policy outperforms two other policies with a significantly lower error rate. The advantage of local-optimal policy over the local-suboptimal policy provides an empirical proof for the \emph{Naive Assumption} in Equation \eqref{navie_assumption} and its monotonicity guarantee in Equation \eqref{navie_assumption_2}.

\subsection{Robust Neural Architecture Search}
\begin{table}[h]
\centering
\caption{Error rates (\%) on CIFAR-10 in different search spaces. The results of DARTS and RobustDARTS are from \cite{DBLP:conf/iclr/ZelaESMBH20}. The best results are highlighted in \textbf{bold}.}
\label{tab:robust_search}
\begin{adjustbox}{max width=\linewidth}
\begin{tabular}{c|ccc}\hline
\textbf{Search Space}  & \textbf{DARTS} & \textbf{RobustDARTS(L2)} & \textbf{CMAB-NAS}        \\ \hline \hline
 S2  &  4.85 & 3.31  & \textbf{2.84 $\pm$ 0.17} \\
 S4  & 7.20 &   \textbf{3.56} & 3.78 $\pm$ 0.27     \\  \hline
\end{tabular}
\end{adjustbox}
\end{table}

\par
It has been shown that the DARTS methods could overfit to the validation set and leads to poor test performance in certain search spaces \cite{DBLP:conf/iclr/ZelaESMBH20}, and in this case, DARTS and improved variants require additional regularization to avoid overfitting. 
Different from DARTS, as a non-differentiable approach, tree-search methods are less prone to such overfitting issues.
Here, we empirically show that our CMAB-NAS can perform the search robustly in different search spaces. 
Here, we consider the two search spaces S2 and S4 tested in RobustDARTS \cite{DBLP:conf/iclr/ZelaESMBH20}, where the original DARTS method demonstrated a clear overfitting issue.
In particular, S2 consists of \{$3 \times 3$ \emph{SepConv}, \emph{SkipConnect}\}, while S4 consist of \{$3 \times 3$ \emph{SepConv}, \emph{Noise}\}. 
We performed the search 3 times with different random seeds under the same hyperparameter setting as in our previous CIFAR-10 experiment.
As shown in Table \ref{tab:robust_search}, our method does not suffer from the overfitting problem of DARTS and works reasonably well even without additional regularizing techniques like the L2 regularization (on the inner objective of DARTS) used in RobustDARTS \cite{DBLP:conf/iclr/ZelaESMBH20}.

\section{Conclusion}
\par
In this paper, we formulated the neural architecture search (NAS) problem as a CMAB problem (CMAB-NAS), which naturally leads us to propose the use of the Nested Monte-Carlo Search (NMCS) with UCB and Top-1 selection policies to solve its two objectives (i.e. \emph{cumulative regret} and \emph{simple regret}). Our CMAB-NAS formulation provides a unified view of current tree-search methods and their efficiency issues.
On CIFAR-10 dataset, our approach discovers a cell structure that can achieve 2.58\% error rate using only 0.58 GPU days, which is 20 times faster than the current state-of-the-art tree-search method. The cell structures discovered by our method on CIFAR-10 transfer well to large-scale dataset like ImageNet.
Our work not only provides a new formulation for NAS but also improves tree-search methods to a performance level that is as effective and efficient as reinforcement learning or differentiable architecture search methods.
This opens up more opportunities for tree-search NAS methods as they are robust to different search spaces and can be continuously improved by exploring more advanced sampling strategies under our CMAB-NAS framework.

\section{Acknowledgement}
This research was undertaken using the LIEF HPC-GPGPU Facility hosted at the University of Melbourne. This Facility was established with the assistance of LIEF Grant LE170100200.

\bibliography{main}

\begin{thebibliography}{10}

\bibitem{imagenet_cvpr09}
J.~Deng, W.~Dong, R.~Socher, L.-J. Li, K.~Li, and L.~Fei-Fei, ``{ImageNet: A
  Large-Scale Hierarchical Image Database},'' in {\em CVPR}, 2009.

\bibitem{krizhevsky2009learning}
A.~Krizhevsky, G.~Hinton, {\em et~al.}, ``Learning multiple layers of features
  from tiny images,'' 2009.

\bibitem{ms_coco_2014}
T.~Lin, M.~Maire, S.~J. Belongie, J.~Hays, P.~Perona, D.~Ramanan,
  P.~Doll{\'{a}}r, and C.~L. Zitnick, ``Microsoft {COCO:} common objects in
  context,'' in {\em ECCV}, 2014.

\bibitem{zoph2017iclr}
B.~Zoph and Q.~V. Le, ``Neural architecture search with reinforcement
  learning,'' in {\em ICLR}, 2017.

\bibitem{zoph2018learning}
B.~Zoph, V.~Vasudevan, J.~Shlens, and Q.~V. Le, ``Learning transferable
  architectures for scalable image recognition,'' in {\em CVPR}, 2018.

\bibitem{pham2018efficient}
H.~Pham, M.~Y. Guan, B.~Zoph, and Q.~V.~L. andf Jeff~Dean, ``Efficient neural
  architecture search via parameter sharing,'' in {\em ICML}, 2018.

\bibitem{sutton2000policy}
R.~S. Sutton, D.~A. McAllester, S.~P. Singh, and Y.~Mansour, ``Policy gradient
  methods for reinforcement learning with function approximation,'' in {\em
  NeurIPS}, 1999.

\bibitem{real2019regularized}
E.~Real, A.~Aggarwal, Y.~Huang, and Q.~V. Le, ``Regularized evolution for image
  classifier architecture search,'' in {\em AAAI}, 2019.

\bibitem{DBLP:conf/gecco/LuWBDDGB19}
Z.~Lu, I.~Whalen, V.~Boddeti, Y.~D. Dhebar, K.~Deb, E.~D. Goodman, and
  W.~Banzhaf, ``Nsga-net: neural architecture search using multi-objective
  genetic algorithm,'' in {\em GECCO}, 2019.

\bibitem{DBLP:journals/corr/abs-2001-01233}
D.~Zhou, X.~Zhou, W.~Zhang, C.~C. Loy, S.~Yi, X.~Zhang, and W.~Ouyang,
  ``Econas: Finding proxies for economical neural architecture search,'' in
  {\em CVPR}, 2020.

\bibitem{DBLP:conf/iclr/ZelaESMBH20}
A.~Zela, T.~Elsken, T.~Saikia, Y.~Marrakchi, T.~Brox, and F.~Hutter,
  ``Understanding and robustifying differentiable architecture search,'' in
  {\em ICLR}, 2020.

\bibitem{negrinho2017deeparchitect}
R.~Negrinho and G.~Gordon, ``Deeparchitect: Automatically designing and
  training deep architectures,'' {\em arXiv:1704.08792}, 2017.

\bibitem{wistuba2017finding}
M.~Wistuba, ``Finding competitive network architectures within a day using
  uct,'' {\em arXiv:1712.07420}, 2017.

\bibitem{liu2018progressive}
C.~Liu, B.~Zoph, M.~Neumann, J.~Shlens, W.~Hua, L.~Li, L.~Fei{-}Fei, A.~L.
  Yuille, J.~Huang, and K.~Murphy, ``Progressive neural architecture search,''
  in {\em ECCV}, 2018.

\bibitem{wang2019alphax}
L.~Wang, Y.~Zhao, Y.~Jinnai, Y.~Tian, and R.~Fonseca, ``Alphax: exploring
  neural architectures with deep neural networks and monte carlo tree search,''
  {\em arXiv:1903.11059}, 2019.

\bibitem{liu2018darts}
H.~Liu, K.~Simonyan, and Y.~Yang, ``{DARTS:} differentiable architecture
  search,'' in {\em ICLR}, 2019.

\bibitem{Chen2019pdarts}
X.~Chen, L.~Xie, J.~Wu, and Q.~Tian, ``Progressive differentiable architecture
  search: Bridging the depth gap between search and evaluation,'' in {\em
  ICCV}, 2019.

\bibitem{chu2019fair}
X.~Chu, T.~Zhou, B.~Zhang, and J.~Li, ``Fair darts: Eliminating unfair
  advantages in differentiable architecture search,'' {\em arXiv:1911.12126},
  2019.

\bibitem{MiLeNAS}
C.~He, H.~Ye, L.~Shen, and T.~Zhang, ``Milenas: Efficient neural architecture
  search via mixed-level reformulation,'' in {\em CVPR}, 2020.

\bibitem{DBLP:conf/iclr/XuX0CQ0X20}
Y.~Xu, L.~Xie, X.~Zhang, X.~Chen, G.~Qi, Q.~Tian, and H.~Xiong, ``{PC-DARTS:}
  partial channel connections for memory-efficient architecture search,'' in
  {\em ICLR}, 2020.

\bibitem{li2020adapting}
Y.~Li, Y.~Wang, C.~Xu, {\em et~al.}, ``Adapting neural architectures between
  domains,'' {\em NeurIPS}, vol.~33, 2020.

\bibitem{cai2018proxylessnas}
H.~Cai, L.~Zhu, and S.~Han, ``Proxyless{NAS}: Direct neural architecture search
  on target task and hardware,'' in {\em ICLR}, 2019.

\bibitem{wan2020fbnetv2}
A.~Wan, X.~Dai, P.~Zhang, Z.~He, Y.~Tian, S.~Xie, B.~Wu, M.~Yu, T.~Xu, K.~Chen,
  {\em et~al.}, ``Fbnetv2: Differentiable neural architecture search for
  spatial and channel dimensions,'' in {\em CVPR}, 2020.

\bibitem{fang2020densely}
J.~Fang, Y.~Sun, Q.~Zhang, Y.~Li, W.~Liu, and X.~Wang, ``Densely connected
  search space for more flexible neural architecture search,'' in {\em CVPR},
  2020.

\bibitem{Mei2020AtomNAS}
J.~Mei, Y.~Li, X.~Lian, X.~Jin, L.~Yang, A.~Yuille, and J.~Yang, ``Atomnas:
  Fine-grained end-to-end neural architecture search,'' in {\em ICLR}, 2020.

\bibitem{wu2020firefly}
L.~Wu, B.~Liu, P.~Stone, and Q.~Liu, ``Firefly neural architecture descent: a
  general approach for growing neural networks,'' {\em NeurIPS}, vol.~33, 2020.

\bibitem{Shu2020Understanding}
Y.~Shu, W.~Wang, and S.~Cai, ``Understanding architectures learnt by cell-based
  neural architecture search,'' in {\em ICLR}, 2020.

\bibitem{zhou2020theory}
P.~Zhou, C.~Xiong, R.~Socher, and S.~C. Hoi, ``Theory-inspired path-regularized
  differential network architecture search,'' {\em NeurIPS}, 2020.

\bibitem{chen2020anti}
H.~Chen, B.~Zhang, S.~Xue, X.~Gong, H.~Liu, R.~Ji, and D.~Doermann,
  ``Anti-bandit neural architecture search for model defense,'' in {\em ECCV},
  pp.~70--85, 2020.

\bibitem{chen2013combinatorial}
W.~Chen, Y.~Wang, and Y.~Yuan, ``Combinatorial multi-armed bandit: General
  framework and applications,'' in {\em ICML}, pp.~151--159, 2013.

\bibitem{ontanon2013combinatorial}
S.~Onta{\~{n}}{\'{o}}n, ``The combinatorial multi-armed bandit problem and its
  application to real-time strategy games,'' in {\em AAAI}, 2013.

\bibitem{cazenave2009nested}
T.~Cazenave, ``Nested monte-carlo search,'' in {\em IJCAI}, 2009.

\bibitem{kocsis2006bandit}
L.~Kocsis and C.~Szepesv{\'{a}}ri, ``Bandit based monte-carlo planning,'' in
  {\em ECML}, 2006.

\bibitem{auer2002finite}
P.~Auer, N.~Cesa{-}Bianchi, and P.~Fischer, ``Finite-time analysis of the
  multiarmed bandit problem,'' {\em Machine Learning}, 2002.

\bibitem{huang2017densely}
G.~Huang, Z.~Liu, L.~van~der Maaten, and K.~Q. Weinberger, ``Densely connected
  convolutional networks,'' in {\em CVPR}, 2017.

\bibitem{SNAS}
S.~Xie, H.~Zheng, C.~Liu, and L.~Lin, ``{SNAS:} stochastic neural architecture
  search,'' in {\em ICLR}, 2019.

\bibitem{inception}
C.~Szegedy, W.~Liu, Y.~Jia, P.~Sermanet, S.~E. Reed, D.~Anguelov, D.~Erhan,
  V.~Vanhoucke, and A.~Rabinovich, ``Going deeper with convolutions,'' in {\em
  CVPR}, 2015.

\bibitem{mobilenet}
A.~G. Howard, M.~Zhu, B.~Chen, D.~Kalenichenko, W.~Wang, T.~Weyand,
  M.~Andreetto, and H.~Adam, ``Mobilenets: Efficient convolutional neural
  networks for mobile vision applications,'' {\em arXiv:1704.04861}, 2017.

\bibitem{loshchilov2016sgdr}
I.~Loshchilov and F.~Hutter, ``{SGDR:} stochastic gradient descent with warm
  restarts,'' in {\em ICLR}, 2017.

\bibitem{devries2017improved}
T.~DeVries and G.~W. Taylor, ``Improved regularization of convolutional neural
  networks with cutout,'' {\em arXiv:1708.04552}, 2017.

\end{thebibliography}
\bibliographystyle{ieeetr}

\appendix
\section{The Best Performing Cell Discovered by our CMAB-NAS}\label{sec:cell}

\begin{figure}[H]
	\centering
	\begin{subfigure}{0.48\linewidth}
		\includegraphics[width=\textwidth]{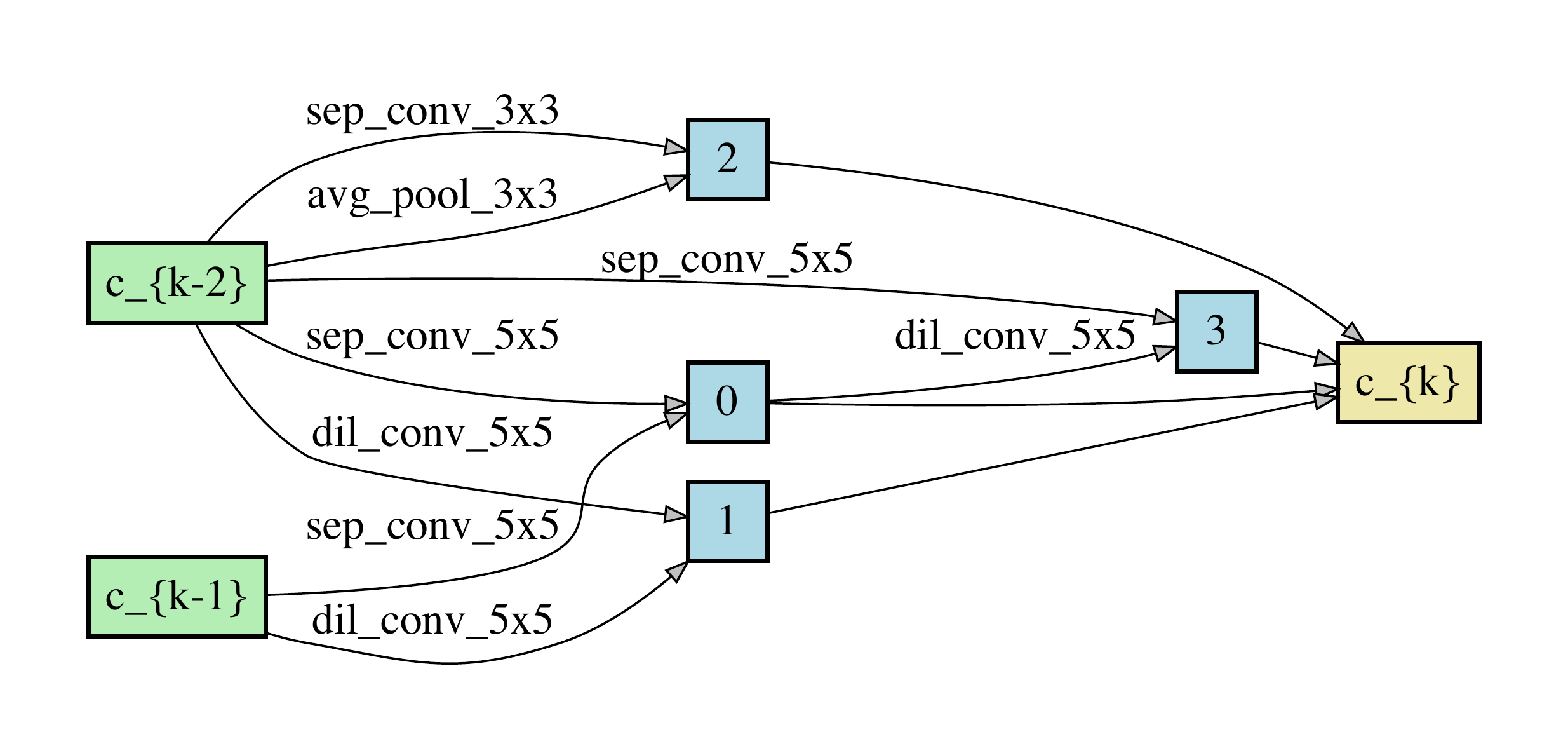}
        \caption{Normal Cell.}
		\label{fig:run3_normal}
	\end{subfigure}
	\begin{subfigure}{0.48\linewidth} 
		\includegraphics[width=\textwidth]{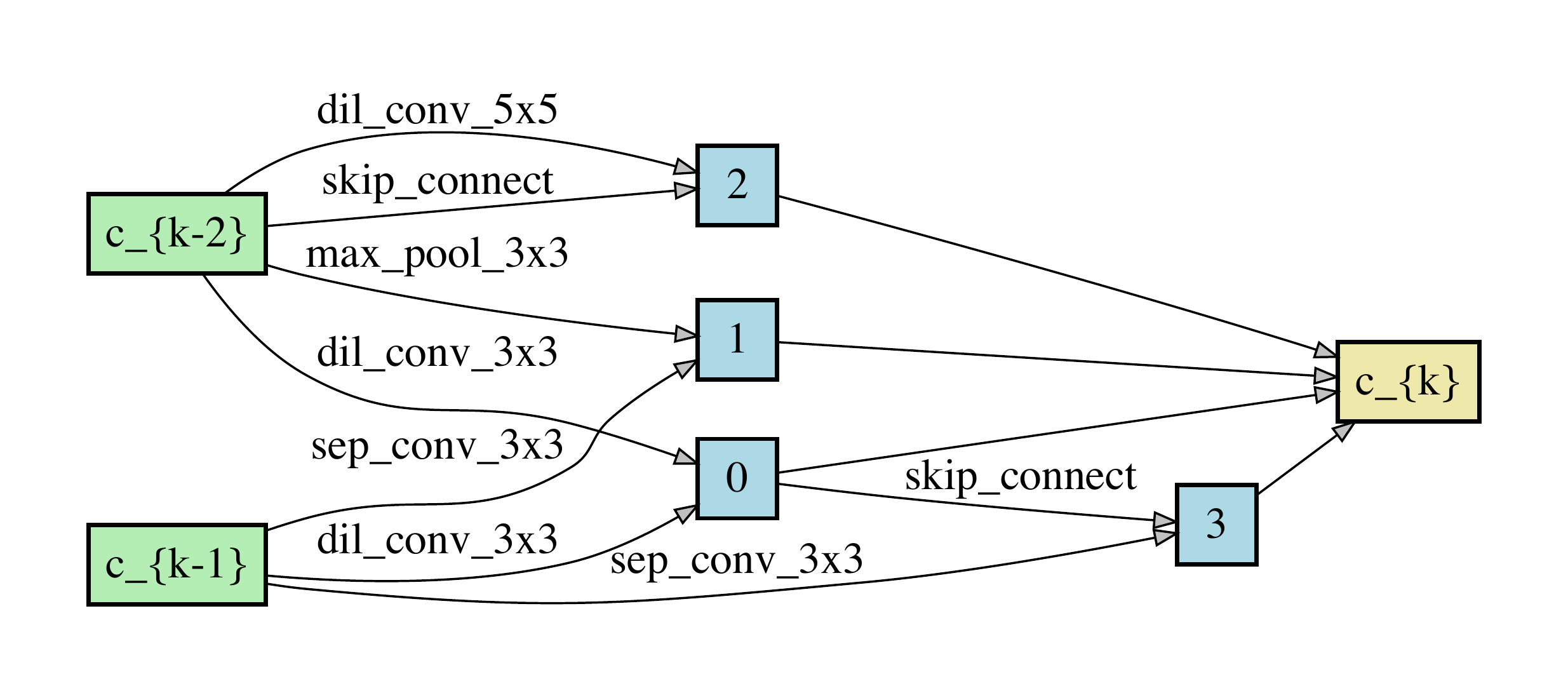}
        \caption{Reduction Cell.}
		\label{fig:run3_reduce}
	\end{subfigure}
	\caption{The best performing cell discovered by our CMAB-NAS}
\end{figure}

\end{document}